\documentclass[11pt]{article}

\usepackage[]{naacl2021}
\usepackage{times}
\usepackage{latexsym}

\usepackage{url}
\usepackage{graphicx}
\usepackage{epstopdf}
\usepackage{amsmath}
\usepackage{color}
\usepackage[labelfont=bf]{caption}
\usepackage{placeins}
\usepackage{wrapfig}
\usepackage{sidecap}
\usepackage{booktabs}
\usepackage{wrapfig}
\usepackage{amssymb}
\usepackage{multirow}
\usepackage{amsthm}
\usepackage{soul}
\usepackage{microtype}
\usepackage{cleveref}
\usepackage{mathtools}
\usepackage[normalem]{ulem}

\newcommand{\pz}{\hphantom{0}}
\newcommand{\pzz}{\hphantom{00}}
\crefformat{section}{\S#2#1#3} % see manual of cleveref, section 8.2.1
\crefformat{subsection}{\S#2#1#3}
\crefformat{subsubsection}{\S#2#1#3}

\title{An Empirical Study of Extrapolation in Text Generation\\ with Scalar Control}
\author{Aashi Jain \\
  UC San Diego \\
  \texttt{a2jain@ucsd.edu} \\
  \And
  Taylor Berg-Kirkpatrick \\
  UC San Diego \\
  \texttt{tberg@ucsd.edu} \\}
\date{}

\begin{document}
\maketitle
\begin{abstract}
We conduct an empirical evaluation of extrapolation performance when conditioning on scalar control inputs like desired output length, desired edit from an input sentence, and desired sentiment across three text generation tasks. Specifically, we examine a zero-shot setting where models are asked to generalize to ranges of control values not seen during training. We focus on evaluating popular embedding methods for scalar inputs, including both learnable and sinusoidal embeddings, as well as simpler approaches. Surprisingly, our findings indicate that the simplest strategy of using scalar inputs directly, without further encoding, most reliably allows for successful extrapolation.
% -- contrary to expectations set in prior work where extrapolation was not directly evaluated.
\end{abstract}

\section{Introduction}
\begin{figure}[t]
    \centering
    \includegraphics[width=\columnwidth]{"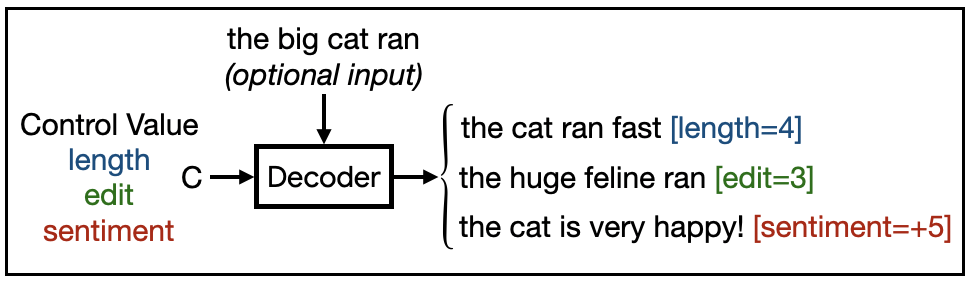"}
    \vspace{-0.3in}
    \caption{Depiction of scalar control tasks.}
    \vspace{-0.2in}
    \label{fig:model_overview}
\end{figure}

Recent work has explored a variety text generation tasks that condition on a control variable to specify a desired trait of the output. Examples include summarization conditioned on a desired output length \citep{fan-etal-2018-controllable}, paraphrase generation conditioned on a parse tree \citep{krishna-etal-2020-reformulating}, style transfer conditioned on sentiment \citep{he2020probabilistic}, and more \citep{hu2017toward, li2018delete, Fu2018StyleTI, He2020LearningSP}.  In this work, we specifically focus on text generation tasks that condition on a \textit{scalar} control variable, as depicted in Figure \ref{fig:model_overview}. 
% which are typically represented using a learnable embedding or a sinusoidal encoding. 
Past work has demonstrated that the embedding strategies used to encode scalars can have substantial impact on control performance \citep{kikuchi2016controlling}. Here, we focus on a novel zero-shot evaluation: how does the embedding method affect the model's ability to generalize from a range of control values seen during training (e.g. sentence lengths $\le$ 10) to a range not seen during training (e.g. sentence lengths $>$ 10)? Empirical knowledge of best practices for enabling generalization to new control values may be useful to the community due to its potential implications for zero-shot tasks in transfer learning and multitask NLP.

% We carry out a comprehensive evaluation on the extrapolation performance of existing and other simpler strategies. Our experiments reveal that embedding techniques from prior work learn to control generation only for the control values that are observed during training\jh{citation?}. Whereas, a fairly simple technique of directly using scalar control values as inputs is surprisingly effective in extrapolating to unobserved control values.

% Specifically, we focus on models that achieve controllable generation by supplementing the decoder inputs with scalar values. Scalar control values are typically represented using $d-$dimensional embedding for each control value. 

% We find that all previous efforts suffered with a common pitfall; it is necessary for the model to observe all potential control values during training.  We believe it is desirable to learn robust models which demonstrate generalization to new control values as they could potentially allow zero-shot learning in transfer and multitask learning. Our experiments find that a simple strategy of using scalar-values directly are better suited for generalization in controlled text generation. 
% This result is especially surprising because it is a common practice to represent scalar values by encoding them before feeding to the neural network\jh{citation}. 

% However, our findings indicate that scalar inputs when used directly as inputs are better suited for generalization in controlled text generation.

To this end, we conduct experiments on both unconditional and conditional text generation across three different tasks: generating from an unconditional language model to match (1) desired output length and (2) desired sentiment, and (3) generating from a conditional paraphrase model to match a desired edit from the input sentence. We compare several different scalar input representation strategies, including learnable embeddings, sinusoidal embeddings, the scalar value itself, as well as vectors of duplicates of the scalar. We evaluate performance using both LSTM \citep{hochreiter1997long} and Transformer \citep{vaswani2017attention} models. We train models on a range of \textit{observed} control values, and then report results on an unobserved extrapolation range. Our results indicate that the simple scalar input control strategy best extrapolates to unobserved values, whereas other strategies extrapolate poorly. This result is especially surprising given that it is relatively a common practice to represent scalar values as vector embeddings before passing them to a neural network \citep{kikuchi2016controlling}. Finally, in our comparison we find that LSTM decoders generalize to unseen control values better than Transformers.

% To this end, we compare different input control strategies and show that scalar input strategy consistently allows generalization to unobserved control values. We conduct experiments to control three tasks: (i)  output length, (ii) divergence between input-output, and (iii) sentiment of the generated text. We train the model on a range of observed control values. We then extrapolate to values both smaller and greater than the control values in the observed range. We conclude that scalar input control strategy successfully extrapolates to unobserved values, whereas other strategies do not. We also show the difference in control behavior exhibited by Long-Short Term Memory (LSTM), \citep{hochreiter1997long} and Transformer, \citep{vaswani2017attention} based decoder models. We find that LSTM decoders generalize to unseen control values better than Transformer decoders.

\section{Related Work}

\citet{kikuchi2016controlling} and~\citet{ liu2018controlling} modified the initial state of the decoder in seq2seq models in order to condition on control values. \citet{sennrich2016controlling} provided control values as an additional input to the encoder instead. Through our initial experiments, we found that the additional inputs to the decoder performed better than other variants.

Controllable generation spans a variety of tasks in language generation. \citet{li2016persona} controlled persona in a conversation model. \citet{mikolov2012context} trained a topic-conditional language model. \citet{hidey2019fixed} controlled semantic edits in the task of arugment generation. \citet{hotate2019controlling} corrected grammatical errors using the word error rate as the control variable.

Apart from the architectural variations, previous work has also experimented with different scalar embedding strategies. \citet{kikuchi2016controlling,takeno2017controlling} used learnable embeddings for each control value as inputs. \citep{takase2019positional} used sinusoidal encodings instead. \citet{lakew2019controlling, ficler2017controlling} experimented with fixed embeddings. \citet{angerbauer2019automatic} encoded each scalar with a one-hot vector. In contrast, we show comparisons with the learnable and sinusoidal strategies for a new task not considered in past work: zero-shot extrapolation. 

\section{Controllable Generation with Scalar Inputs}
\label{define_task_model}
Following past work on scalar inputs for controlled text generation, the models considered in our empirical investigation have a common form. For sequence to sequence tasks, the goal is to learn a conditional probability $P(Y|X,c)$, where $X, Y$ are the input and output text respectively, and $c$ is the additional scalar control variable. Similarly, to add additional scalar constraints to a language model, we learn the probability distribution conditioned only on the scalar control values, $P(X|c)$.

\subsection{Tasks}
We investigate the generalization performance of scalar controllable text generation on three tasks. In all the tasks we have considered in this work, the scalar inputs are integer valued. 

\vspace{5pt} \noindent \textbf{Length Control:} We control the length of the output text in two settings: unconditional and conditional text generation. For unconditional text generation, we train a language model with an additional length constraint. At test time, the language model is used to sample text of desired length. In the Appendix \ref{sec:append_additional_results}, we also report results to control length for a paraphrase generation task. Successful generalization to unobserved lengths can be useful for transfer learning tasks -- for example, length control in an abstractive summarization model can be used for long-form text generation. 

\vspace{5pt} \noindent \textbf{Edit/Divergence Control:} For the task of paraphrase generation, it is also desirable to control the edit of output from the input. Further allowing paraphrase models to generalize to edit/divergence levels not seen during training could allow for various use cases, including data augmentation methods with superior diversity. 
We use the Jaccard Distance to control this divergence of the output text from the input. We compute the desired edit value by multiplying the jaccard distance with $10$ and rounding to the nearest integer. Using this procedure, we obtain edit values in range $[0-10]$. 

\vspace{5pt} \noindent \textbf{Sentiment Control:} Sentiment controlled generation aims to generate text with a desired sentiment attribute. Being able to generalize to unobserved value might allow such models to be trained from text data with more limited sentiment range, opening up further use cases. Unlike our other tasks, we expect the difficulty of sentiment extrapolation to be predominantly  driven by unseen lexical items like words or bigrams with extremely high sentiment values. 

\begin{figure}
    \centering
    \includegraphics[scale=0.45]{"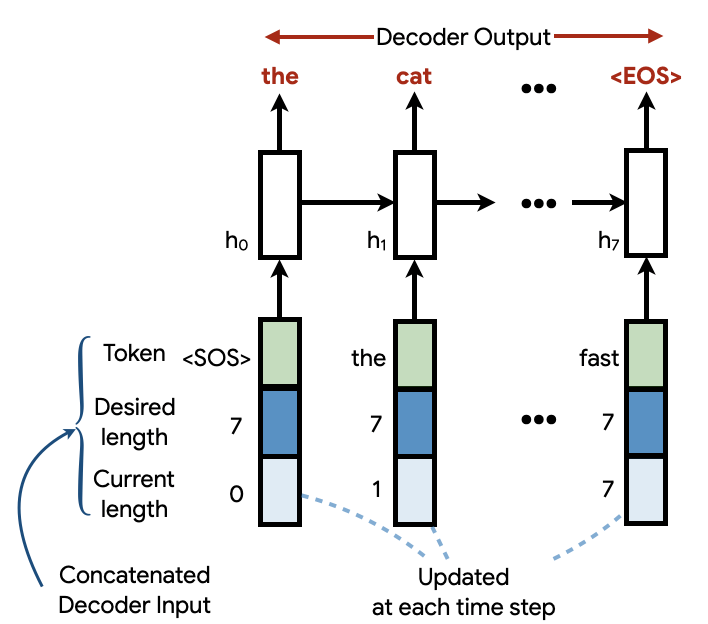"}
    \caption{Decoder architecture to control scalar attributes like length (and edit). At each time step, the decoder's input token embedding is concatenated with the desired control value embedding and the current output's control value embedding.}
    \label{fig:model}
    \vspace{-0.2in}
\end{figure}

\subsection{Control Architecture}
Following existing literature \citep{kikuchi2016controlling}, we pass control values as an additional input to the decoder, as depicted in Figure \ref{fig:model}. We concatenate the scalar control embedding with the input token embedding of the decoder. For some of the scalar inputs we investigate, like length and edit, we add an additional embedding that keeps track of the current length or edit of the current output from the input. This additional embedding provides the model with information that tracks how well decoding has matched the control value so far, at each time step.

\section{Scalar Embedding Strategies}
In this section, we describe the different embedding strategies. These input strategies are then coupled with model architectures explained earlier to obtain a controllable text generation model.

\subsection{Learnable Embedding}
In the first control strategy we study, the  model learns an embedding matrix for the potential control values it can encounter during training \citep{takeno2017controlling}. Learnable embedding gives the model the flexibility to learn weights for each scalar control independently. However, the need to train an embedding from scratch for each control value makes it an unsuitable candidate for extrapolation and we expect it to perform poorly. 

\subsection{Sinusoidal Encoding}
Next, we experiment with the sinusoidal encoding which was introduced for the transformer models \citep{vaswani2017attention} for encoding the notion of order in a sequence. The encoding is formed by projecting the scalar input onto a sequence of periodic basis functions with different frequencies. In principle, as long as the period of the lowest frequency basis functions extend past our extrapolation range, sinusoidal embeddings could allow the model to generalize to unobserved values.

\subsection{Direct Scalar Value}
Lastly, we investigate a simple strategy of passing the scalar control value itself to the model. Instead of creating a fixed-size embedding, we directly use the scalar value as an input to the decoder.

Further, we consider a final variant of this approach: we form an embedding by duplicating the scalar control value $d$ times to create a $d$-dimensional vector, which we name scalar\_repeat. Like other embeddings, these encodings are then concatenated with the decoder's input token embedding. Since neural training objectives are non-convex and duplicate scalar values will each touch different model parameters in the input layer, we expect this strategy may lead to different outcomes than the single scalar version. More specifically, the duplicated strategy effectively gets multiple chances at having input layer parameters randomly initialize to useful values that leads to a productive local optimum. 

% We believe that repetition of the scalar input could help draw more focus on the information provided by the scalar inputs.
\begin{table*}[t]
\small
\centering
\scalebox{0.9}{
    \begin{tabular}{cccccccccc}
        \toprule
         \multicolumn{2}{c}{Dataset} & \multicolumn{3}{c}{SNLI}& \multicolumn{2}{c}{Quora}& \multicolumn{3}{c}{Yelp}\\
         \multicolumn{2}{c}{Trained}& \multicolumn{3}{c}{Observed $L \leq 20$} & \multicolumn{2}{c}{Observed $E \leq 7$} & \multicolumn{3}{c}{Observed $S \geq 2$ }    \\
         \cmidrule(l){1-2}  \cmidrule(l){3-5} \cmidrule(l){6-7} \cmidrule(l){8-10}
         
         \multicolumn{2}{c}{Evaluated}& $L \leq 30$ & \multicolumn{2}{c}{$ 20<L \leq 30$} &  $E \leq 10$ &  $E > 7$ & $S \geq 2$ &  \multicolumn{2}{c}{$S<2$}    \\
         
        \cmidrule(l){1-2} \cmidrule(l){3-3} \cmidrule(l){4-5} \cmidrule(l){6-6} \cmidrule(l){7-7} \cmidrule(l){8-8} \cmidrule(l){9-10}
         
        Model & Strategy   & PPL $\downarrow$ & Acc. $\uparrow$ & MSE $\downarrow$  & PPL $\downarrow$ & MSE $\downarrow$ & PPL $\downarrow$ & Acc. $\uparrow$  & MSE $\downarrow$\\
         \midrule
        \multirow{5}{*}{LSTM} & no\_control & 24.4  & -- & --& 13.9   & --  & 42.3     & -- & -- \\
         & learnable & 31.4  & \pz10.0 & 196.3       & 20.1   & 23.6 & \textbf{42.1}      & \pz1.0 & 15.7     \\
         & sinusoidal& 24.1  & \pzz8.9 & \pz39.9 & 12.2   & 12.3 & 42.3      & \textbf{37.0} & \pz4.1    \\
         & scalar    & \textbf{19.8}  & \pz93.3 & \pzz0.1 & \textbf{11.4}   & \pz7.1 & \textbf{42.1}      & 20.0 & \pz2.7      \\
         & scalar\_repeat       & 23.2 & \textbf{100.0} & \textbf{\pzz0.0} & 11.9   & \textbf{\pz4.6} &  43.1      & 35.0 & \textbf{\pz1.7}      \\
         \midrule
    \multirow{5}{*}{Transformer} & no\_control & 23.4  & -- & --& 11.1   & --  & \textbf{38.2}      & -- & -- \\
         & learnable & 24.8  & \pz11.1 & 175.8       & 10.6   & 21.4 & 38.3      & \pz4.0 & 14.7     \\
         & sinusoidal& 21.1  & \pz13.3 & \pz72.0 & \textbf{\pz9.8 }   & 20.5 & 38.3      & 15.0 & \pz5.6     \\
         & scalar    & \textbf{18.4}  & \textbf{\pz62.2} & \textbf{\pzz0.5} & \textbf{\pz9.8}    & 21.7 & -- & -- & -- \\
         & scalar\_repeat       & 18.7  & \pzz7.8 & \pz33.9 & 10.9   & \textbf{10.1} & 38.5      & \textbf{36.0} & \textbf{\pz2.8 }     \\
         \bottomrule
    \end{tabular}}
\caption{Comparison of five input control strategies on three datasets. Observed range indicates control values used to train the model. Evaluated range depicts the controls values for which PPL, Accuracy, and MSE are reported on test set. $L$, $E$, and $S$ stand for Length, Edit, and Sentiment respectively.}
\label{tab:results}
\vspace{-0.1in}
\end{table*}

\section{Experimental Setup}

\subsection{Decoder Models}
In our experiments, we train using both LSTM and Transformer models. For LSTM decoders, the scalar embedding is concatenated with the decoder input at each time step. Similarly for the Transformer models, we concatenate additional scalar embedding to the token embedding for all decoder inputs before feeding to the decoder layers.

\subsection{Datasets}
We conduct experiments on three datasets. To train a length-controlled language model we subsample examples from the SNLI dataset \citep{bowman2015large} for lengths $\leq 30$. For edit controlled generation, the Quora Question Pairs (QQP) \footnote[1]{https://www.kaggle.com/c/quora-question-pairs/data}, a popular paraphrase generation dataset is used. We also report additional results on length control on this dataset in the Appendix \ref{sec:append_additional_results}. For sentiment control, we use the Yelp dataset \citep{tang2015document} with review ratings on the scale of $1-5$.

\subsection{Training and Evaluation}
For all the tasks and datasets, perplexities (PPL) are reported on the test set. We also report BLEU scores on the Quora dataset. These values are reported to test the quality of the generated text. To evaluate the effectiveness of the model at generating controlled text, we first train the model on a range of observed scalar inputs. Then, we evaluate on both the observed and unobserved range of values. We report the Mean Squared Error (MSE) between the desired scalar input and control value of the generated text on both these intervals. We also report Accuracy that calculates the percentage of examples for which the model generates the text with the exact desired control value. To measure accuracy on sentiment control task, we separately train a classifier on the Yelp dataset. Then the classifier's predictions are used to label the sentiment of the generated text. Full experiment details can be found in Appendix~\ref{sec:appendix}.

\begin{figure}[t]
    \centering
    \includegraphics[width=.38\textwidth]{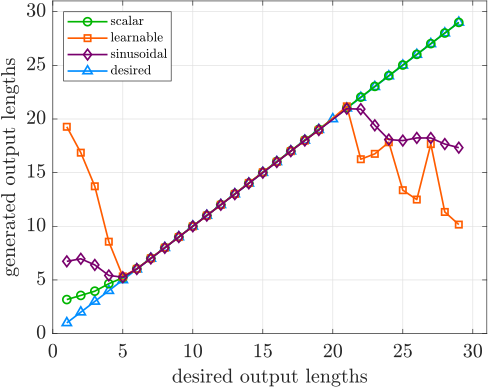}
    \caption{Generated output lengths for different input control strategies on the SNLI dataset. Lengths $L\leq 20$ represent the observed values used to train the model.}
    \label{fig:plot}
    \vspace{-0.2in}
\end{figure}
\section{Results}

In Table \ref{tab:results}, we report results on the unobserved control values for all tasks and models. We also report PPL for the \textit{no\_control} case, where we model uncontrolled generation. For each task, we report the control values on which it was trained, noted in the \textit{Trained} row. L $\leq$ 20 for the SNLI task means that model was trained on lengths less than or equal to 20. Similarly, control ranges for other tasks are defined in the table. The tasks are evaluated on different intervals reported by \textit{Evaluated} row.

Comparing the five control strategies, we find that \textit{scalar} and \textit{scalar\_repeat} generally achieve lower MSE values across the three tasks. Especially, for the SNLI task to control length using LSTM, both the \textit{scalar} and \textit{scalar\_repeat} strategies result in near perfect scores on Accuracy and MSE. We also find that \textit{scalar} scored the lowest perplexity across all tasks and models. 

For the most part, Accuracy and MSE tend to agree with each other. However, this does not hold for sentiment control on Yelp dataset using the LSTM model. Even though the \textit{sinusoidal} strategy achieves a higher accuracy, its MSE is worse compared to \textit{scalar} and \textit{scalar\_repeat}. This indicates that when \textit{sinusoidal} gets the sentiment wrong, it gets it wrong by a high margin. Whereas, \textit{scalar} and \textit{scalar\_repeat} both generate text closer to the desired sentiment on average. 

We do not report results on \textit{scalar} for the transformer sentiment control because of the limitations introduced by the hyperparameters chosen for the Transformer architecture. For more details, we direct the reader to the Appendix \ref{sec:scalar_yelp}. \textit{Scalar\_repeat} results in better Accuracy and MSE compared to \textit{sinusoidal} indicating superior generalization.

Overall, \textit{scalar} and \textit{scalar\_repeat} generalize to unobserved values better than other strategies. Figure \ref{fig:plot} supports this observation, we see that \textit{scalar} continues the upward trend to unobserved lengths $L>20$. Moreover, \textit{scalar} perfectly aligns with the desired output larger lengths even in the unobserved range. It should be noted even though the model was trained on lengths $L\leq20$, the dataset did not have text for lengths $L<5$. 
% This also reflects that both \textit{learnable} and \textit{sinusoidal} fail to respect the downward trend beyond the observed range. Whereas, \textit{scalar} generalizes better on these smaller unobserved lengths than other strategies as well. 
% \jh{To describe more clearly as a conclusion, you can just say other baselines fail to extrapolate lengths well while the proposed method generalizes to unseen lengths successfully.}

\section{Conclusion}

In this work, we carried out an investigation for controlled text generation with different input embedding strategies conditioned on length, edit, and sentiment scalar inputs. We compared learnable, sinusoidal, and direct scalar input strategies across three text generation tasks. We found that the simple strategy of providing scalar inputs directly to the decoder generalizes surprisingly well to unobserved control values. These results might serve as a guide for researchers focusing on related zero-shot settings in future work.
\bibliography{custom}
\bibliographystyle{acl_natbib}
\appendix

\section{Appendix}
\subsection{Details of the Experimental Setup}
\label{sec:appendix}
For constructing the SNLI dataset, we randomly sample text with length $\leq30$, consisting of 100K/10K/10K for training, validation, and test set. For constructing the Quora Question Pairs (QQP) dataset, we split the dataset into training, validation, and test set such that no split shares a common question. We also filter out examples with output lengths greater than 30 from all the splits. The trainiing, validation, and test split consists of roughly 115K/4K/30K examples respectively. The Yelp sentiment dataset is constructed by sampling text with lengths $\leq 100$. We create balanced splits, where each split contains roughly the same text for all sentiment ratings. The splits training, validation, and test contains 100K/10K/10K examples respectively. To train the Yelp sentiment classifier, we sample about 1 million text from the dataset using a Transformer model. We achieve an accuracy of about $65\%$ on the test set. For tokenization, in each task we make use of the Spacy tokenizer. We generate text using greedy decoding for seq2seq models. For sampling text from the language model, we employ temperature sampling.

\begin{table*}[t]
\small
\centering
\scalebox{1}{
\begin{tabular}{p{0.1\linewidth}p{0.9\linewidth}}
    \toprule
    Length & Generated Output   \\
    \midrule
    % 21 & a person with no clothes and a black shirt is standing around being area in front of a white wall .  \\
    % 25 & a woman wearing a brown helmet and a pink snowboard laying down outside by the table and looks out from a piece of steps .  \\
    13 & a boy sleeping in his driveway going to arrive on the beach . \\
    18 & a man takes a boat in the urban water looking at the fathers market behind a destination .\\
    19 & a man in a group of students , standing on a bench during a white lit device . . \\
    \midrule
    Sentiment & Generated Output   \\
    \midrule
    % 5& good food , fast service , clean bathrooms and large atmosphere . i will be back , that 's ! ! \\
    % 5& good burgers and tater tots are bomb ! ! ! price is average and service is great ! i usually have a soda but i ca n't complain on the food and for no matter how much you 're going to do the food is awesome ! ! ! \\
    5 & its a friendly , ill get you the best curry in town .   the panang curry is wonderful ! yum   i tried a korean curry so i ca nt rate it with 3 stars tonight i thought it was great since i went to chow mein for dinner .   was inside the noodle restaurant outside and service were great and the staff was attentive .   would like to give her a try during lunch time and they need to adjust my schedule . \\
    5 & nice place to grab a smoothie drinks to go . they are super friendly and they have great tap choices if you 're in a rush .\\
    5 & we really liked the food .   and the service has been great with reservations in the valley ... so i would go back .   their biscuits are usually hard to find .   it 's not money brunch !   food is excellent !   nice menu !\\
    \bottomrule
\end{tabular}}
\caption{Generated outputs by the \textit{scalar} control model for unobserved lengths and sentiment.}
\label{table:generations}
% \vspace{-0.2in}
\end{table*}

\subsection{Justification for \textit{scalar} results on Yelp}
\label{sec:scalar_yelp}
In this section, we describe the reason the results for \textit{scalar} are not reported in the experimental results. To train a conditional text generation model for sentiment on Yelp dataset, we set $n\_heads=3$. The embedding size of the input tokens is set to $256$ for all control strategies. Concatenating the desired scalar sentiment to the embedding results the input size not divisible by $n\_heads$, a necessary constraint imposed by the model architecture. Instead, we only report results on $scalar\_repeat$ strategy models on Yelp dataset.
\subsection{Generated Outputs}
In Table \ref{table:generations}, we report cherry-picked output generations for all tasks. For text generated by controlling sentiment, the model was trained on sentiment $S \leq 3$. Therefore, the model never really trains on any positive, $S=4$ or $S=5$, sentiment text. It is interesting to see that it learns to generate positive text. We also show text generated by conditioning on length, where the observed lengths are $\leq 10$.

\subsection{Additional Quantitative Results}
\label{sec:append_additional_results}
In this section, we report additional results for each task described in the work. Table \ref{tab:snli_len} reports the results for controlling length on the SNLI dataset for five input control strategies. We train models on two splits of training data, (i) for lengths $7\leq L \leq 30$ and (ii) for lengths $L\leq 20$. For each of the training splits, we include results for observed range (same range on which the model was trained) and the unobserved range of control values.

In Table \ref{tab:quora_len}, we report results for controlling length on the Quora Question Pairs dataset. The training and evaluation splits are similar to the SNLI length splits. Please refer to the previous paragraph for more information.

Results to control edit on the Quora Question Pairs dataset are reported in Table \ref{tab:quora_edit}. We use the Jaccard Distance used to measure the edit between input and output text. As described in the paper, we multiply the Jaccard Distance by 10 and round the obtained metric to the nearest integer. This procedure results in edit values in the range $[0-10]$. Similar to other tasks, we train on two different splits of the data: (i) $E \geq 3$ and (ii) $E \leq 6$. For each training split, we further report values on the observed edit values and the unobserved edit values. 

Lastly, results for controlling sentiment on the Yelp dataset are reported in Table \ref{tab:yelp}. The Yelp dataset contains text rated on a scale of $[1-5]$. The two splits of the training data are created: (i) $S \geq 2$ and (ii) $S\leq3$. 
% SNLI LEN TABLE
\begin{table*}[t]
\small
\centering
\scalebox{0.95}{
    \begin{tabular}{cccccccccccc}
    \toprule
   \multicolumn{2}{c}{Trained} & \multicolumn{5}{c}{Observed $7 \leq L \leq 30$}  & \multicolumn{5}{c}{Observed $L \leq 20$}  \\
   \cmidrule(l){1-2} \cmidrule(l){3-7} \cmidrule(l){8-12}
   
    \multicolumn{2}{c}{Evaluated} & $L \leq 30$ & \multicolumn{2}{c}{ $7 \leq L \leq 30$ } & \multicolumn{2}{c}{$L < 7$} & $L \leq 30$ & \multicolumn{2}{c}{$L \leq 20$} & \multicolumn{2}{c}{ $20<L \leq 30$} \\
    
    \cmidrule(l){1-2} \cmidrule(l){3-3} \cmidrule(l){4-5}  \cmidrule(l){6-7} \cmidrule(l){8-8} \cmidrule(l){9-10} \cmidrule(l){11-12} 
    Model & Strategy & PPL  $\downarrow$ & Acc. $\uparrow$ & MSE  $\downarrow$ & Acc. $\uparrow$ & MSE $\downarrow$ & PPL  $\downarrow$  & Acc. $\uparrow$ & MSE  $\downarrow$ & Acc. $\uparrow$ & MSE $\downarrow$ \\
     \midrule
  \multirow{4}{*}{LSTM}    & no\_control  & 26.6 & --     & --   & --   & --  & 24.4  & --     & --    & --   & --  \\
     & learnable  & 27.3 & \pz68.3  & 4.9 & \pz0.0 & 152.7 & 31.4 & 52.5   & 18.8 & \pz10.0 & 196.3\\
     & sinusoidal & 24.9 & \pz59.6  & 0.7 & \pz0.0 &\pz41.1 & 24.1 & 43.0   & \pz0.7  & \pzz8.9& \pz39.9 \\
     & scalar  & 22.0 & \pz98.7 & 0.0 & 15.0 & \pzz3.6 & 19.8 & \textbf{95.5}   & \textbf{\pz0.0}  & \pz93.3 & \pzz0.1 \\
     & copy\_scalar  & 23.8 & \textbf{100.0} & \textbf{0.0}  & \textbf{55.0} & \textbf{\pzz0.8} & 23.2 & \textbf{95.5} & \textbf{\pz0.0} & \textbf{100.0} & \textbf{\pzz0.0} \\
    \midrule
  \multirow{4}{*}{Transformer}& no\_control  & 25.1 & --     & --   & --   & --  & 23.4 & --     & --    & --   & --  \\
     & learnable  & 26.8 & 82.6  & 3.1 & \pz0.0 & 247.1 & 24.8 & \textbf{82.0}   & 22.4 & 11.1 & 175.8 \\
     & sinusoidal & 20.4 & \textbf{87.4}  & \textbf{0.1} & \pz1.7 & \pz26.5 & 21.1 & 80.0   & \pz1.3 & 13.3 & \pz72.0 \\
     & scalar  & 21.2 & 48.7   & 0.6 & \pz0.0 & \pz18.8 & 18.4 & 81.0   & \textbf{\pz1.2} & \textbf{62.2} & \textbf{\pzz0.5} \\
     & copy\_scalar  & 22.7 & 35.7 & 1.9 & \textbf{\pz8.3} &   \textbf{\pzz3.2}  & 18.7 & 49.5   & \pz1.5 & \pz7.8& \pz33.9 \\
     \bottomrule
    \end{tabular}
}
\caption{Results on the SNLI dataset for length controlled text generation.}
\label{tab:snli_len}
\vspace{-0.2in}
\end{table*}

% QUORA LEN TABLE
\begin{table*}[t]
\small
\centering
\scalebox{0.85}{
    \begin{tabular}{cccccccccccccc}
        \toprule
        \multicolumn{2}{c}{Trained} & \multicolumn{6}{c}{Observed $7 \leq L \leq 30$}    & \multicolumn{6}{c}{Observed $L \leq 20$}    \\
        \cmidrule(l){1-2} \cmidrule(l){3-8} \cmidrule(l){9-14}
        \multicolumn{2}{c}{Evaluated}  & \multicolumn{2}{c}{$1 \leq L \leq 30$}  & \multicolumn{2}{c}{ $7 \leq L \leq 30$} & \multicolumn{2}{c}{$L < 7$} & \multicolumn{2}{c}{$1 \leq L \leq 30$} & \multicolumn{2}{c}{$L \leq 20$} & \multicolumn{2}{c}{$20 $<$ L \leq 30$} \\
         
        \cmidrule(l){1-2} \cmidrule(l){3-4} \cmidrule(l){5-6} \cmidrule(l){7-8} \cmidrule(l){9-10} \cmidrule(l){11-12} \cmidrule(l){13-14}
         Model & Strategy & PPL $\downarrow$ & BLEU $\uparrow$ & Acc. $\uparrow$ & MSE $\downarrow$ & Acc. $\uparrow$ & MSE $\downarrow$ & PPL $\downarrow$ & BLEU $\uparrow$ & Acc. $\uparrow$ & MSE $\downarrow$  & Acc. $\uparrow$ & MSE $\downarrow$ \\
        \midrule
        \multirow{4}{*}{lstm}   & no\_control & 11.0 & 18.2 & --     & --   & -- & -- & 11.8 & 18.1 & --     & --    & -- & --\\
         & learnable & 10.8 & 17.0 & 91.8  & 9.5 & \pz0.0\pz & 217.3 & 18.7 & 17.0 & 75.4  & 36.1 & 10.1 & 133.0 \\
         & sinusoidal& \pz9.6 & 18.6 & 88.2  & 0.1 & \pz0.0 & \pz71.9 & 14.4 & 18.3 & 75.5  & \pz3.9 & 10.2 & \pz53.5 \\
         & scalar     & \pz9.3 & 20.3 & \textbf{95.3}  & \textbf{0.1} & \textbf{10.0} & \textbf{\pzz4.5} & \pz9.8 & 18.3 & \textbf{79.2}  & \textbf{\pz0.5} & \textbf{94.9}& \textbf{\pzz0.1} \\
        \midrule
        \multirow{4}{*}{tf} & no\_control & \pz8.7 & 23.6 & --     & --   & -- & --& \pz9.2 & 22.6  & --     & --    & -- & --\\
         & learnable & \pz8.1 & 22.5 & \textbf{65.4}  & 4.6 & \pz0.0 & 144.2 & 14.4 & 21.2  & \textbf{60.5}   & 26.9 & \textbf{7.8} & 177.0\\
         & sinusoidal& \pz7.3 & 23.4 & 52.3  & \textbf{0.7} & \pz0.0 & \pz29.9 & \pz9.8 & 21.6  & 47.1  & \textbf{\pz3.6}  & 6.4 & \pz60.2 \\
         & scalar  & \pz7.8 & 23.1  & 44.6  & 1.0 & \pz0.1 & \pz27.9 & \pz9.7 & 22.0 & 14.7  & \pz6.6  & 1.1 & \pz37.7 \\
         & scalar\_repeat  & \pz8.7 & 24.9 & 25.8 & 4.7 & \textbf{25.7} & \textbf{\pzz2.9} & \pz8.2 & 21.9 & 27.9 & 10.3 & 5.0 & \textbf{\pz34.7\pz} \\
        \bottomrule 
    \end{tabular}
}
\caption{Results on the Quora Question Pairs dataset for length controlled paraphrase generation.}
\label{tab:quora_len}
\vspace{-0.2in}
\end{table*}

% QUORA EDIT TABLE
\begin{table*}[t]
\small
\centering
\scalebox{1}{
\begin{tabular}{ccrrrrrrrr}
\toprule
 \multicolumn{2}{c}{Trained} & \multicolumn{4}{c}{Observed $E \leq 7$} & \multicolumn{4}{c}{Observed $E \geq 4$} \\
 \cmidrule(l){1-2} \cmidrule(l){3-6} \cmidrule(l){7-10}
 \multicolumn{2}{c}{Evaluated} & \multicolumn{2}{c}{$0 \leq E \leq 10$} & $E \leq 7$ & $E > 7$ & \multicolumn{2}{c}{$0 \leq E \leq 10$} & $E \geq 4$ & $E <4$  \\
  
 \cmidrule(l){1-2} \cmidrule(l){3-4} \cmidrule(l){5-5} \cmidrule(l){6-6} \cmidrule(l){7-8} \cmidrule(l){9-9} \cmidrule(l){10-10}
Model  & Strategy   & PPL $\downarrow$ & BLEU $\uparrow$ & MSE $\downarrow$ & MSE $\downarrow$ & PPL $\downarrow$ & BLEU $\uparrow$ & MSE $\downarrow$ & MSE $\downarrow$ \\

  \midrule
\multirow{5}{*}{LSTM} & no\_control& 13.9 & 20.5 & \multicolumn{1}{c}{--}   & \multicolumn{1}{c}{--}     & 13.5 & 16.4 & \multicolumn{1}{c}{--}  & \multicolumn{1}{c}{--}     \\
 & learnable& 20.1& 19.0 & 3.2 & 23.6 & 13.2 & 15.3 & 3.5 & 19.0 \\
 & sinusoidal     & 12.2 & \textbf{20.9} & 3.5 & 12.2 & 11.9 & 16.3 & 3.9 & 11.3 \\
 & scalar    & \textbf{11.4} & 20.6 & 3.3 & 7.1  & \textbf{11.7} & \textbf{18.2} & 3.7 & 7.8   \\
\multicolumn{1}{l}{}  & \multicolumn{1}{l}{scalar\_repeat} & 12.2 & 20.0 & \textbf{3.0} & \textbf{4.6}  & 12.2 & 17.1  & \textbf{3.1} & \textbf{7.4}  \\
\midrule
\multirow{5}{*}{Transformer}     & no\_control& 11.1 & 23.7 & \multicolumn{1}{c}{--}  & \multicolumn{1}{c}{--}   & 10.5 & 20.3 & \multicolumn{1}{c}{--} & \multicolumn{1}{c}{--}     \\
 & learnable& 10.6 & 23.5 & 3.6 & 21.4 & 9.4 & \textbf{21.1} & \textbf{4.1} & 11.0    \\
 & sinusoidal     & \textbf{9.8} & 24.0 & 4.5 & 20.5 & \textbf{9.2} & 20.8 & 7.0 & 7.8  \\
 & scalar    & \textbf{9.8} & 24.0 & 5.2 & 21.7  & 9.3 & 20.4 & 10.7 & 6.8 \\
  & scalar\_repeat & 10.9 & \textbf{25.6} & \textbf{4.1} & \textbf{10.1} & 11.3 & 20.1 & 5.7 & \textbf{6.7} \\
\bottomrule
\end{tabular}}
\caption{Results on the QQP dataset for divergence controlled paraphrase generation.}
\label{tab:quora_edit}
\vspace{-0.2in}
\end{table*}

% YELP SENTIMENT TABLE
\begin{table*}[th]
\small
\centering
\scalebox{1}{
    \begin{tabular}{cccccccccccc}
        \toprule
        \multicolumn{2}{c}{Trained} & \multicolumn{5}{c}{Observed $2 \leq S \leq 5$}& \multicolumn{5}{c}{Observed $1 \leq S \leq 3$} \\
        \cmidrule(l){1-2} \cmidrule(l){3-7} \cmidrule(l){8-12}
        \multicolumn{2}{c}{Evaluated} & \multicolumn{3}{c}{$2 \leq S \leq 5$} & \multicolumn{2}{c}{$S < 2$} &  \multicolumn{3}{c}{$1 \leq S \leq 3$} & \multicolumn{2}{c}{$3 < S$} \\
        \cmidrule(l){1-2} \cmidrule(l){3-5} \cmidrule(l){6-7} \cmidrule(l){8-10} \cmidrule(l){11-12}
        Model & Strategy & PPL $\downarrow$ & Acc. $\uparrow$ & MSE $\downarrow$ & Acc. $\uparrow$ & MSE $\downarrow$ & PPL $\downarrow$ & Acc. $\uparrow$ & MSE $\downarrow$ & Acc. $\uparrow$  & MSE $\downarrow$ \\
        \midrule
        \multirow{5}{*}{LSTM} & no\_control & 42.3 & -- & --  & -- & --&  & -- & --  & --  & --\\
        & learnable & 42.1& \textbf{49} & 1.0 & 01 & 15.8 & 44.7  & \textbf{54} & 0.8 & 14  & 2.2 \\
        & sinusoidal& 42.3 & 41 & 1.0 & \textbf{37} & \pz4.1  & 44.9  & 49 & 0.9 & 12  & 4.0 \\
        & scalar & 42.1 & 41 & 1.1 & 20 & \pz2.7 & 44.9  & 49 & 0.8 & 16 & \textbf{2.2} \\
        & scalar\_repeat & 43.1 & 46 & \textbf{1.0} & 35 & \textbf{\pz1.7} & 44.9  & 52 & \textbf{0.7} & \textbf{20}  & 2.2 \\
        \midrule
        \multirow{5}{*}{Transformer} & no\_control & 38.2 & -- & --  & -- & --&  & -- & --  & --  & --\\
        & learnable & 38.3 & 39 & 1.2 & 4 & 14.7 & 43.1  & 43 & 1.0 & \pz2  & 6.2 \\
        & sinusoidal& 38.3 & 44 & 1.0 & 15 & \pz5.6 & 42.8  & 46 & 1.0 & \textbf{20}  & 2.3 \\
        & scalar\_repeat & 38.5 & \textbf{48} & \textbf{0.9} & \textbf{36} & \textbf{\pz2.8} & 42.7  & \textbf{49} & \textbf{0.9} & 18  & \textbf{1.8} \\
        \bottomrule
    \end{tabular}
}
\caption{Results on the Yelp Reviews Dataset for sentiment controlled generation.}
\label{tab:yelp}
\vspace{-0.2in}
\end{table*}
\end{document}